\documentclass{INTERSPEECH2023}

 \interspeechcameraready

\usepackage[T1]{fontenc}
\usepackage{etoolbox}

\makeatletter
\patchcmd{\maketitle}{\@fnsymbol}{\@alph}{}{}  
\makeatother

\title{Efficient Multimodal Neural Networks for Trigger-less Voice Assistants}
\name{Sai Srujana Buddi$^1$, Utkarsh Oggy Sarawgi$^1$, Tashweena Heeramun$^1$, Karan Sawnhey$^1$, Ed Yanosik$^1$, Saravana Rathinam$^{2*}$\thanks{* Work performed at Apple.}, Saurabh Adya$^1$}
\address{
  $^1$ Apple, USA\\
  $^2$ elbo.ai, USA}
\email{\{sbuddi,usarawgi,t\_heeramun,ksawhney\}@apple.com}

\begin{document}

\maketitle
 
\begin{abstract}
The adoption of multimodal interactions by Voice Assistants (VAs) is growing rapidly to enhance human-computer interactions. Smartwatches have now incorporated trigger-less methods of invoking VAs, such as Raise To Speak (RTS), where the user raises their watch and speaks to VAs without an explicit trigger. Current state-of-the-art RTS systems rely on heuristics and engineered Finite State Machines to fuse gesture and audio data for multimodal decision-making. However, these methods have limitations, including limited adaptability, scalability, and induced human biases. In this work, we propose a neural network based audio-gesture multimodal fusion system that (1) Better understands temporal correlation between audio and gesture data, leading to precise invocations (2) Generalizes to a wide range of environments and scenarios (3) Is lightweight and deployable on low-power devices, such as smartwatches, with quick launch times (4) Improves productivity in asset development processes.
\end{abstract}

\section{Introduction}

Multimodality is a fundamental aspect of human interactions \cite{kurosu2020human, rasenberg2022multimodal}, as speech is frequently combined with other modalities, such as gesture, and gaze \cite{TURK2014189}. Human Computer Interfaces (HCIs) have embraced this concept to enhance natural modes of communication \cite{oviatt2003advances}. However, despite this trend, physical button press (BP) and voice trigger (VT) using keyword detection remain the dominant ways of invoking and interacting with Voice Assistants (VAs) on personal devices \cite{nayak2022improving}. These methods can be cumbersome, particularly in multi-turn conversations \cite{qin2021proximic}. Raise to Speak (RTS), developed by Zhao et al. \cite{zhao2019raise}, address these issues, and provides a more natural and convenient way of interacting with VAs on smartwatches using gesture and speech. The user can simply raise their watch to their mouth and speak, without the need for any explicit trigger phrases. The RTS task presents a complex challenge that requires the integration of multiple modalities:
\begin{itemize}
    \item Recognizing and identifying the combined parallel and sequential pattern of gesture (raise) and speech (audio).
    \item Having the ability to adapt to a wide range of user scenarios, physical movements, demographics, and languages.
    \item Running efficiently on low-power devices such as smartwatches, with limited resources.
    \item Achieving a balance between fast response time and high precision.
\end{itemize}
RTS system described in \cite{zhao2019raise} employs human-designed Finite State Machines (FSMs) for audio-gesture integration, but FSMs have limitations in capturing intricate human behavior accurately, lack adaptability, and are susceptible to human biases. The difficulty in designing FSMs increases as the system complexity grows \cite{harel1987statecharts}. In this work, we present a refined method for solving the RTS problem by incorporating neural networks to seamlessly merge the audio and gesture modalities for improved VA invocation accuracy on smartwatches. We have devised \textit{Neural Policy}, a lightweight, and low latency Gated Recurrent Unit (GRU) \cite{chung2014empirical} that integrates speech and gesture signals and makes a binary RTS trigger/no-trigger decision. Due to the loose coupling between the speech and gesture modalities \cite{1407896, kettebekov2003improving, pouw2019quantifying} and resource constraints of a low-power device like a smartwatch, we adopt a late fusion approach \cite{nagrani2021attention, 9190246}. Our proposed Neural Policy delivers significant accuracy improvements, with a $90\%$ relative reduction in False Reject Rate (FRR) and False Accept Rate (FAR) while also eliminating the complexity in system design process.

\begin{figure}[h]
\centering
\begin{subfigure}{\linewidth}
  \centering
  \includegraphics[width=\linewidth, height=0.45\linewidth]{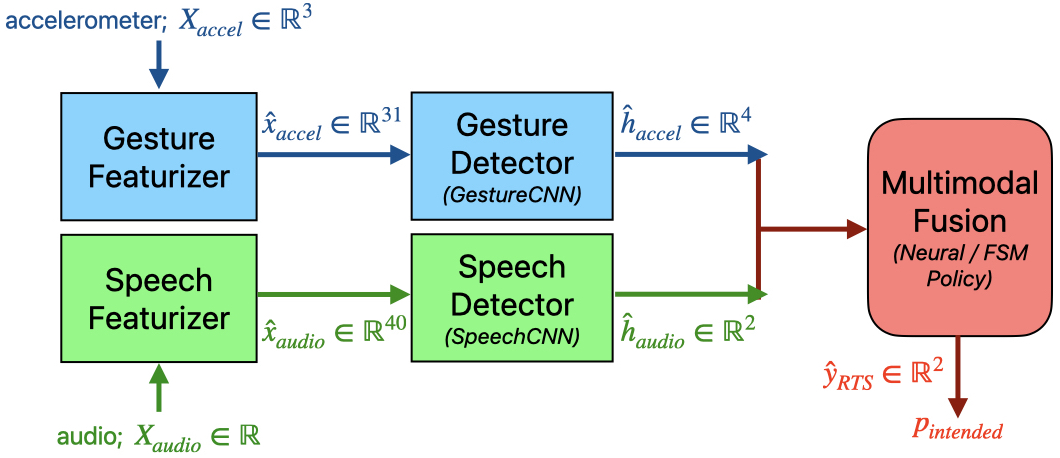}
  \caption{RTS detection model overview.}
  \label{fig:rts_detection}
\end{subfigure}%
\\
\begin{subfigure}{\linewidth}
  \centering
  \includegraphics[width=0.9\linewidth, height=0.6\linewidth]{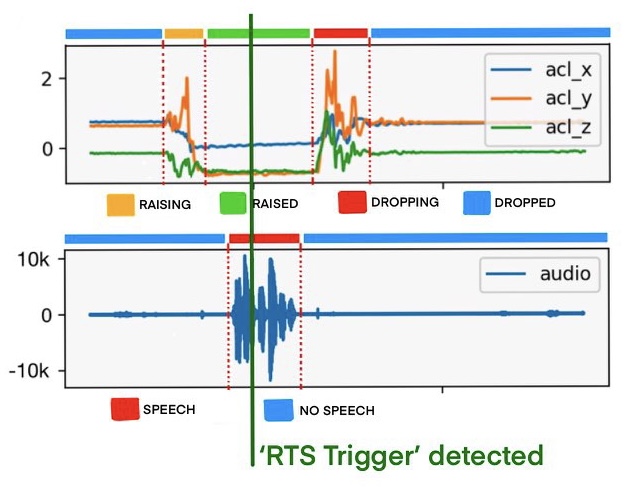}
  \caption{An example timing diagram of an RTS trigger. $acl\_x$, $acl\_y$, $acl\_z$ corresponds to the 3-axis accelerometer values.}
  \label{fig:timing_diagram}
\end{subfigure}
\caption{Raise to Speak (RTS) model and process overview.}
\label{fig:rts_main}
\end{figure}

\section{Related work}
Multimodal learning has a long-standing history that predates and continues through the deep learning era \cite{Ramachandram2017DeepML}. The advancements in deep learning have been instrumental in the considerable growth of multimodal learning \cite{summaira2021recent}, leading to surge of interest in multimodal interfaces, with focus on improving the flexibility, transparency, and expressiveness of HCIs \cite{TURK2014189, ceraso2014re, dumas2009multimodal}. 

Gesture signals have been used in various multimodal applications, such as speech disambiguation \cite{holzapfel2004implementation} and object manipulation in augmented reality \cite{williams2020understanding}. Recently, there has been an increasing trend of using gestures to enhance HCIs \cite{Qiu_2020, brown2007shake2talk, 1389771, 7472168}. Zhao et al. introduced an innovative way of using gesture and speech to make the VA invocation process quick, intuitive, and natural \cite{zhao2019raise}. However, most of the multimodal systems defined in HCI works \cite{zhao2019raise, 1389771, 7472168, 9812670, donkal2018multimodal} use rule-based or linear systems for multimodal fusion, which fail to adapt to a diverse and sizable population. \cite{Qiu_2020, 7169562} explored using deep learning for gesture and speech fusion but these systems are not computationally efficient for low-power devices.

\section{Methods and Materials}
\subsection{RTS Detection: System and Components}\label{sec:model_and_proces}
The goal of the RTS detection systems is to detect RTS triggers by understanding the correlation and underlying patterns of accelerometer and audio signals as shown in Figure \ref{fig:timing_diagram}. To account for the asynchronous nature of input signals and device limitations, we chose a late fusion approach for this system. This approach involves processing each modality separately and then fusing the processed modalities to arrive at a multimodal decision. Figure \ref{fig:rts_detection} shows an overview of RTS detection system, with its components:
\begin{itemize}
    \item Audio featurizer, extracts $40$ dimensional mel filter banks, $\hat{x}_{audio} \in \mathbb{R}^{40}$, from input audio $X_{audio} \in \mathbb{R}$.
    \item Gesture featurizer, extracts $31$ dimensional statistical features \cite{zhao2019raise}, $\hat{x}_{accel} \in \mathbb{R}^{31}$, from $3-$axis accelerometer data $X_{accel} $. 
    \item Speech Detector, a binary streaming classifier for detecting speech using two classes - \textit{speech}, \textit{no speech}. It takes in the audio features $\hat{x}_{audio}$ and outputs $\hat{h}_{audio} \in \mathbb{R}^{2}$.
    \item Gesture Detector, a $4$-way streaming classifier for recognizing the RTS gesture, defined by a valid hand raise, using four classes - \textit{raising}, \textit{raised}, \textit{dropping}, \textit{dropped}. It takes in gesture features $\hat{x}_{accel}$ and outputs $\hat{h}_{accel} \in \mathbb{R}^{4}$.
    \item Multimodal fusion, which aligns the audio and gesture detector models outputs, $\hat{h}_{audio}$ and $\hat{h}_{accel}$, and makes a binary trigger decision, $p_{intended}$, using two classes - \textit{Not Intended for VA} and \textit{Intended for VA}, $\hat{y}_{RTS} \in \mathbb{R}^{2}$. 
\end{itemize}

\subsection{Training setup}\label{sec:training_setup} 
We adopt a sequential training process for development of the RTS system and components discussed in Section \ref{sec:model_and_proces}. Initially, we train the unimodal speech and gesture detectors separately, and the resulting scores from the unimodal detectors are then combined to train the multimodal fusion. This sequential training setup has several benefits, including reducing the complexity of the model and infrastructure, and enabling the utilization of all available data, even when paired modalities or labels are missing. Designing this system under real-world constraints, especially when dealing with a significant amount of missing data and labels in the gesture modality, presented a significant challenge. Nonetheless, the sequential training setup facilitated maximizing the utilization of the available data.

\subsection{Dataset}\label{sec:data}    
In order to gather data that is representative and high in quality, we collected data from smartwatches, specifically, audio recordings through the single-channel $16$ kHz microphone built into the smartwatch, as well as $3-$axis accelerometer data through the smartwatch's accelerometer sensor. To create a meaningful, and diverse dataset, we developed a detailed user study program. Through this program, we collected a total of $8000$ hours of data from $1500$ hired subjects with consent, with each subject recording $100-150$ sessions. We ensured chosen subjects are from a variety of demographics, including gender, ethnicity, and location, as well as physical characteristics like height, age, and preferred wrist for wearing the watch.

We chose an orchestrated approach to collect data, in which a moderator guides subjects through a wide range of scenarios defined by a data collection protocol. The objective of the protocol is to guarantee that the gathered data encompasses the diverse range of dynamic environments and activities that a smartwatch could potentially encounter in real-world settings. During the data collection process, subjects perform activities such as sitting, standing, walking, running, cycling, driving, kayaking, stretching, gesturing, and lying down, in environments including meeting rooms, gyms, parks, crowded areas, and quiet rooms. Half the scenarios involve the user interacting with the VA using RTS. Each subject receives $75$ instructions, each with different activity and environment combinations, and records one or two sessions per instruction. Additionally, we collected video from the moderator's point of view to aid annotation.

The data is split into train, validation, and test sets- with ratios of $70\%$, $15\%$, and $15\%$, respectively. To maintain reliability in the model development process, we ensure that each subject's recordings belong to only one split. The data are annotated with audio, gesture, and intent to interact with VA labels. Audio from the watch is used for audio labeling and video from the recordings are used for gesture and intent labeling.

\subsection{Evaluation Metrics}
Owing to the user experience of the interaction, we perform evaluation of the detection system on the following metrics:
\begin{itemize}
    \item False Reject/Miss Rate (FRR): ratio of triggers that are missed by the detector to the total number of true triggers. A low FRR is desirable. 
    \item False Accept/Wake Rate (FAR): measures the frequency with which the detector is falsely triggered. A low FAR is desirable.
    \item Equal Error Rate (EER): point on Detection Error Tradeoff (DET) curve \cite{martin1997det} where FAR meets FRR. A low EER is preferred.
\end{itemize}
In addition to accuracy metrics, we evaluate on-device performance metrics such as \textit{runtime latency, physical memory, and power consumption}.

\section{Multimodal Temporal Neural Fusion}\label{sec:main}
We propose \textit{Neural Policy}, that uses lightweight, temporal neural networks to fuse audio and gesture signals for improved responsiveness, and enhanced RTS performance. Such a neural fusion technique, is able to use vast amounts of diverse training data to learn patterns, and correlations between the underlying modalities. Our approach involves constructing streaming temporal networks that integrate the outputs from the RTS gesture and speech detectors on a per-frame basis and generate a probability of occurrence of RTS invocation in a streaming fashion.

Firstly, the unimodal Gesture and Speech detectors as detailed in Section \ref{sec:model_and_proces} are trained as $10Hz$ streaming classifiers with a $4$-way and $2$-way classification setup, respectively. Both detectors are trained using Adam optimizer \cite{kingma2014adam} to minimize the frame-wise discriminative categorical cross-entropy loss in accordance with their corresponding frame-wise labels as defined in Equation \eqref{cel} below.
\begin{equation}
  l_{CE} = -\sum_{i=1}^{n}\sum_{j=1}^{m}\sum_{k=1}^{K}y_{k}^{(i,j)}log(\hat{s}_{k}^{(i,j)}) 
  \label{cel}
\end{equation}
\begin{equation}
\text{where softmax scores, } \hat{s}_{k}^{(i,j)} = \frac{e^{\hat{l}_{k}^{(i,j)}}}{\sum_{j=1}^K e^{\hat{l}_{k}^{(i,j)}}}
  \label{softmax}
\end{equation}
$\hat{l}$ represents output of the detector's final layer (logits) , $n$ represents sample count, $m$ represents number of frames per session, and $K$ represents number of classes where $K=2$ for the speech detector, and $K=4$ for the gesture detector. 

\begin{figure}[h]
\centering
\begin{subfigure}{\linewidth}
  \centering
  \includegraphics[width=\linewidth]{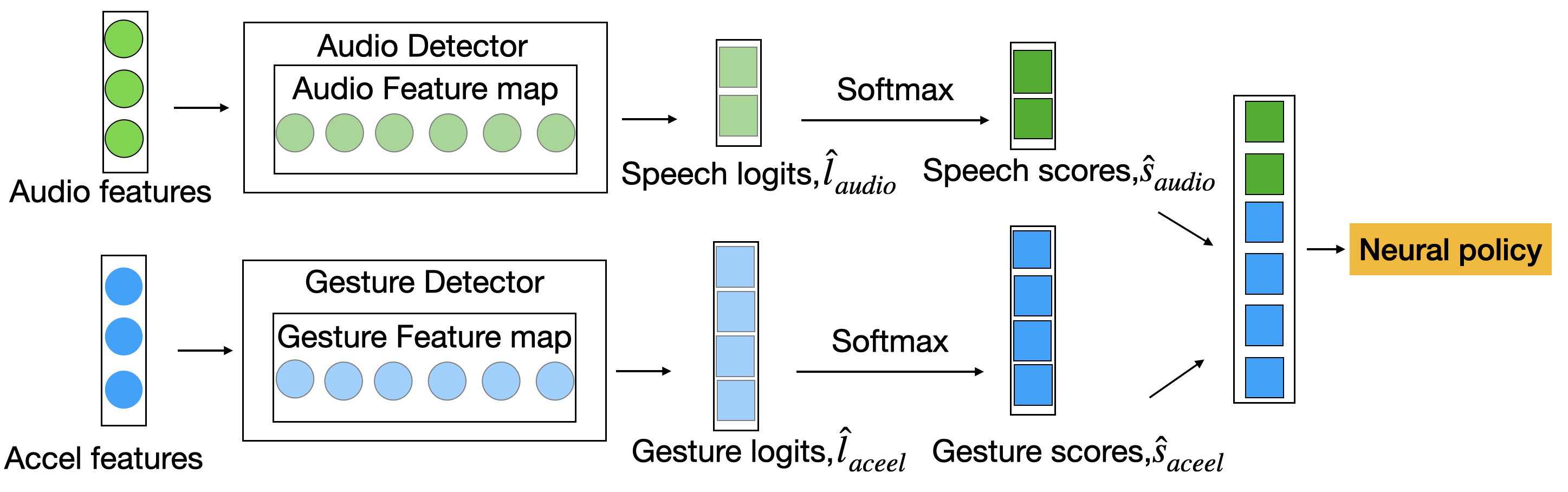}
  \caption{Softmax neural fusion}
  \label{fig:softmax}
\end{subfigure}
\\
\vspace{4mm}
\begin{subfigure}{\linewidth}
    \centering
  \includegraphics[width=\linewidth]{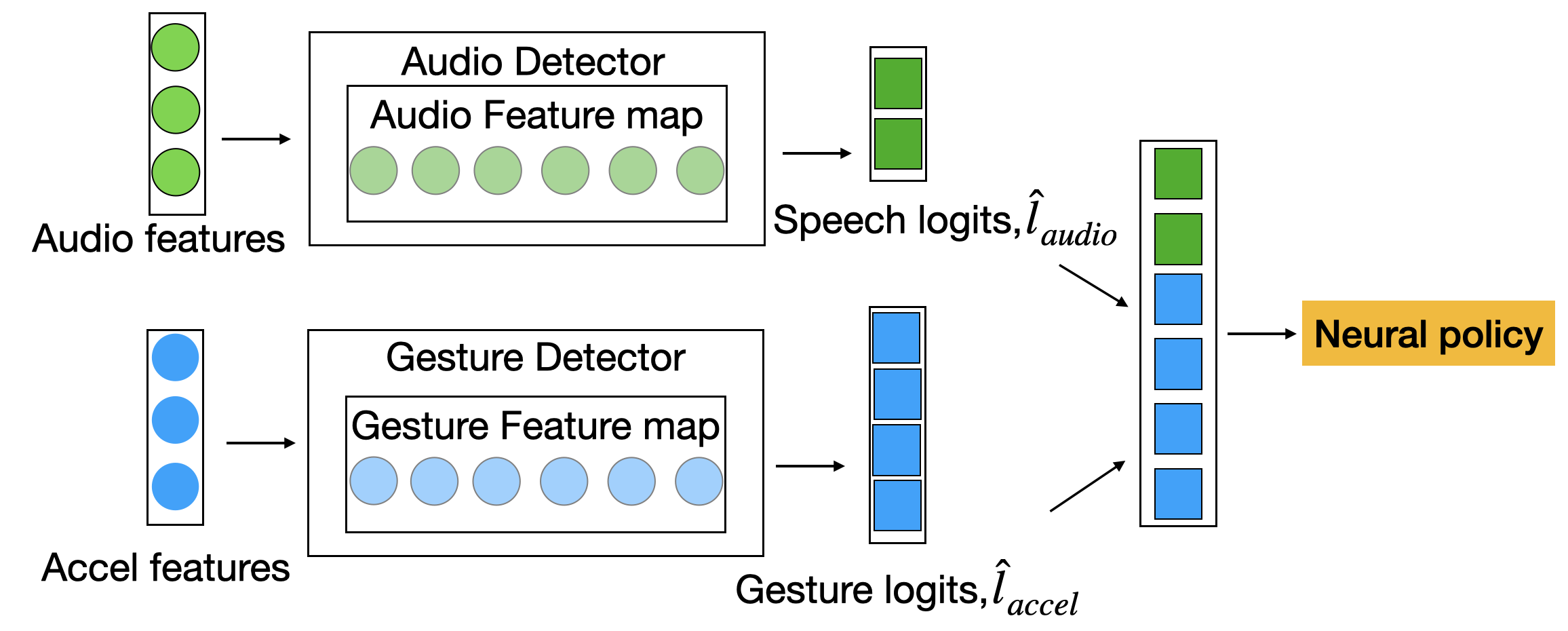}
  \caption{Logit neural fusion}
  \label{fig:logits}
\end{subfigure}
\caption{Types of Neural Fusion}
\label{fig:neural_fusion}
\end{figure}

The outputs from the trained speech and gesture detectors at the frequency of $10Hz$ are aligned, padded, and merged and are used as inputs to develop the neural policy. As part of the neural policy development, we propose two variations of the multimodal neural fusion technique, \textit{softmax neural fusion} and \textit{logit neural fusion}, as also illustrated in Figure \ref{fig:neural_fusion}.

\subsection{Softmax neural fusion}\label{sec:softmax_fusion} 
In softmax neural fusion approach, the frame-wise $4$ dimensional, and $2$ dimensional output probability score sequences from the gesture, and speech detectors respectively are used as input to the neural policy as shown in Figure \ref{fig:softmax}. These output probability scores are computed by applying softmax function to the unimodal detectors final output as defined by Equation \eqref{softmax}. The aligned audio and gesture scores are merged and used to train the neural fusion model using the Adam optimizer \cite{kingma2014adam} to minimize the frame-wise discriminative categorical cross-entropy loss in accordance with their frame-wise binary \textit{intended/unintended} labels defined by Equation \eqref{cel}.

\subsection{Logit neural fusion} \label{sec:logit_fusion} 
Softmax non-linearity pushes the scores towards the extremes of 0 and 1, causing perturbations in the fusion latent space by reducing its dynamic range, thus making raw outputs of models like embeddings and logits more suitable for downstream tasks and further computations \cite{horiguchi2019significance}. Considering this, we propose \textit{logit neural fusion}, which incorporates the $4$ and $2$ dimensional gesture and speech logits directly in the fusion process of neural policy. The final layer output of the unimodal detectors, without applying the softmax function, serve as inputs to the neural policy, as depicted in Figure \ref{fig:logits}. With device limitations in mind, logits are preferred over embeddings. The training procedure of logit neural fusion is similar to that of softmax neural fusion, as discussed in Section \ref{sec:softmax_fusion}. The merged $2$ and $4$ dimensional audio and gesture logits are used to train the fusion model using the Adam optimizer \cite{kingma2014adam} to minimize the frame-wise discriminative categorical cross-entropy loss in accordance with their frame-wise binary \textit{intended/unintended} labels.  

\subsection{Model architecture} \label{sec:model_arch}
Due to their lightweightedness and effectiveness in speech detection and gesture recognition tasks, we chose a $1-$dimensional convolutional neural network (1DCNN) architecture \cite{zhao2019raise, jia2021marblenet, 9751601}, for the speech and gesture detectors. Each 1DCNN detector consists of convolutional layers that operate along the temporal dimension, followed by a flatten, and two dense layers with dimensions $128$ and $32$.

Due to the sequential nature and temporal ordering of the data, we opted Gated Recurrent Units (GRU) for the architecture of \textit{neural policy}. Our chosen GRU architecture consists of a $6$ dimensional input layer, followed by a single hidden layer with $h_{dim}$ dimensions, followed by a fully connected layer, that results in $2$ dimensional output. Softmax function performed on the output results in probability scores for RTS invocation occurrence. Although we have explored Long Short-Term Memory networks (LSTMs) as well for this task, we found that GRUs are more efficient as they require fewer parameters and computational resources to achieve high accuracy.

\section{Experiments and Results}
We experimented with the hyperparameters of the models described in Section \ref{sec:model_arch}, and produced $36$ distinct RTS detectors, by varying the size of the GRU neural policy's hidden layer, $h_{dim} \in \{32, 64\}$, the number of convolutional layers in unimodal 1DCNN detectors, $n \in \{1,3,5\}$, and the type of neural fusion, $f \in \text{\{softmax, logit\}}$. All the models involved are trained over $200$ epochs with a learning rate of $1e^{-3}$. In order to ensure an unbiased comparison, we also implemented an FSM policy as outlined by Zhao et al. \cite{zhao2019raise} using $15$ hyperparameters, and included this in our experimentation, resulting in a total of $54$ distinct variations of the RTS detector. All the 54 variations are trained, tuned and evaluated end-to-end. We have selected a subset of models for further analysis based on their accuracy and on-device performance. Out of the $54$ variations of the RTS detector, $6$ candidates we have hand-picked  for further evaluation are:
\begin{itemize}
    \item(a) : $n_{s}=1$, $n_{g}=1$, $f=FSM$  (baseline)
    \item(b) : $n_{s}=1$, $n_{g}=1$, $f=Softmax$, $h_
{dim}=32$
    \item(c) : $n_{s}=1$, $n_{g}=1$, $f=Softmax$, $h_
{dim}=64$
    \item(d) : $n_{s}=1$, $n_{g}=1$, $f=Logit$, $h_
{dim}=64$
    \item(e) : $n_{s}=3$, $n_{g}=1$, $f=Logit$, $h_
{dim}=64$
    \item(f) : $n_{s}=5$, $n_{g}=1$, $f=Logit$, $h_
{dim}=64$
\end{itemize}
$n_s, n_g$ represents number of 1DCNN layers in speech and gesture detectors respectively, $f$ represents neural fusion type, and $h_{dim}$ represents size of GRU neural policy's hidden layer.

\begin{table}[h]
  \caption{EER comparison of RTS detectors. \#Speech, \#Gesture, and \#NP are number of parameters in Speech, Gesture and Neural Policy detectors respectively.}
  \label{tab:model_comp}
  \centering
  \begin{tabular}{ccccc}
    \toprule
      & \textbf{\#Speech} & \textbf{\#Gesture} & \textbf{\#NP} & \textbf{EER(\%)}\\
       \toprule
       (a) & 133466 & 133356 & 15 & 40.0\\
       (b) & 133466 & 133356 & 3906 & 8.1\\
       (c) & 133466 & 133356 & 13954 & 7.0\\
       (d) & 133466 & 133356 & 13954 & 5.5\\
       (e) & 58546 & 133356 & 13954 & 4.9\\
           \textbf{(f)} & \textbf{77482} & \textbf{133356} & \textbf{13954} & \textbf{4.4}\\
    \bottomrule
  \end{tabular}
\end{table}

\begin{figure}[h]
      \centering
      \includegraphics[width=0.8\linewidth, height=0.8  \linewidth, scale=0.8]{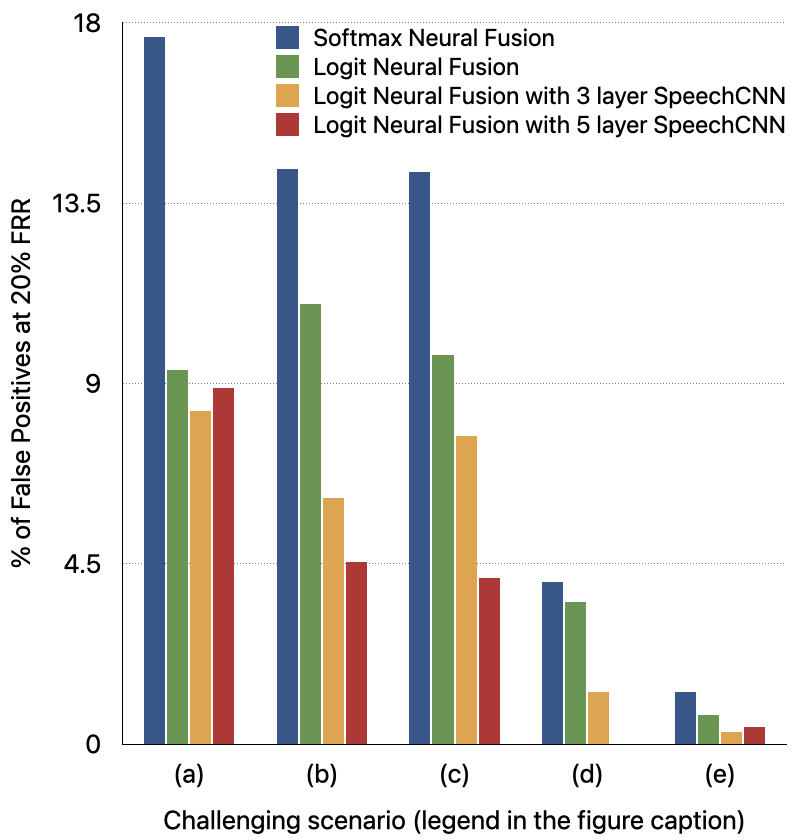}
      \caption{False Positives of challenging scenarios at fixed 20\% FRR and variable FAR. (a) Check time on watch while talking (b) Check  watch and read what’s on it (c) Raise watch near mouth and cough (d) Hold phone in same hand as watch and attend a call (e) Perform a right turn while on steering wheel and speak loudly.}
      \label{fig:challenging_scenarios}
    \end{figure}

Table \ref{tab:model_comp} compares accuracy and model parameters of the selected RTS detectors, and several key observations can be drawn, (1) Replacing FSM with a neural policy using softmax neural fusion results in an $80\%$ relative decrease in EER, indicating that the neural policy approach is highly effective, and the neural networks are more capable of learning from a large population than FSM. It is also observed that increase in the size of the neural policy leads to further accuracy improvements. (2) The accuracy of neural fusion is further enhanced by using logit neural fusion, which addresses the limitations of softmax non-linearity, as discussed in Section \ref{sec:logit_fusion}. (3) Increasing the depth of the speech detector and adding more 1DCNN layers enhances the audio representation, leading to neural policy accuracy improvements. Figure \ref{fig:challenging_scenarios} illustrates that RTS detectors with deeper speech detectors have higher accuracy in challenging negative scenarios, where RTS gestures are paired with unintended human speech. However, increasing the depth of the gesture detector did not lead to any performance improvements, likely due to the relatively smaller dataset available for gesture. The RTS detector that employs a 5-layered speech detector, a 1-layered gesture detector, and a logit neural policy with hidden size 64 (highlighted in Table \ref{tab:model_comp}) demonstrates the most significant accuracy gains and achieves a relative reduction in EER of $90\%$.

\setlength{\tabcolsep}{4pt} 
\renewcommand{\arraystretch}{1} 
\begin{table}[h]
  \caption{On-device performance metrics of RTS detectors. The values are recorded by running detectors on 32 bit ARM Cortex-A7 MPCore processor in low power mode, averaged over 30 iterations. Memory is physical memory consumed in megabytes. Power measures static and dynamic power drawn per inference along with static power of device when idle in milliWatts. Runtime is time taken per inference measured in milliseconds.}
  \label{tab:perf}
  \centering
  \begin{tabular}{ccccccc}
    \toprule
    & \textbf{(a)} & \textbf{(b)} & \textbf{(c)} & \textbf{(d)} & \textbf{(e)} & \textbf{(f)} \\
    \midrule
    \textbf{Memory(MB)} & 1.04 & 1.06 & 1.06 & 1.06 & 0.77 & 0.85\\
    \textbf{Power(mW)} & 100.4 & 122.2 & 122.4 & 122.4 & 124.1 & 128.9\\
    \textbf{Runtime(ms)} & 0.30 & 0.54 & 0.57 & 0.57 & 0.76 & 1.4\\
    \bottomrule
  \end{tabular}
\end{table}

    Table \ref{tab:perf} compares the on-device performance metrics for the selected RTS detectors. The results are competitive, demonstrating that the chosen Neural Policy, despite using neural networks, is lightweight and suitable for on-device implementation. In this analysis, the models precision is set to FP32, but we observed that quantizing them to FP16 yields even better competitive outcomes with no accuracy reductions.
    
    The neural policy not only improves accuracy but also significantly simplifies the asset development process. The FSM policy development which involves no training, but finetuning in a $15$ dimensional hyperparameter space to find a single operating point, takes $3-5$ days and $1000$ compute nodes even with the efficiency of Bayesian search and parallelism. In contrast, training and tuning the neural policy using distributed data parallel on a single $4$ GPU node takes $2-3$ hours reducing development time and compute resources significantly.
        
\section{Conclusion and Future Work}
In conclusion, this work introduces two versions of a GRU-based multimodal temporal neural fusion approach, one using softmax and the other using logit, for accurate identification of VA trigger-less invocations using audio and gesture signals. The approach demonstrates its ability to learn the temporal correlations between the input audio and gesture signals and achieve high precision. It is lightweight and suitable for deployment on low-power devices, such as smartwatches, while maintaining low launch time and latency. Compared to the traditional FSM-based multimodal fusion, this approach results in a significant improvement in the Raise To Speak experience, with a relative decrease of nearly $90\%$ in false rejects and false accepts. Additionally, this approach offers a more streamlined development process when compared to the traditional FSM-based multimodal fusion.

Future research efforts can be focused on experimenting with techniques such as modality dropout \cite{neverova2015moddrop} to make the fusion more robust. Attention-based fusion models like multimodal transformers \cite{xu2022multimodal} and attention bottlenecks \cite{nagrani2021attention} can be explored, albeit it is crucial to ensure that performance metrics such as memory, latency, and power are kept within acceptable limits for on-device applications.




\bibliographystyle{IEEEtran}
\bibliography{mybib}

\begin{thebibliography}{10}
\providecommand{\url}[1]{#1}
\csname url@samestyle\endcsname
\providecommand{\newblock}{\relax}
\providecommand{\bibinfo}[2]{#2}
\providecommand{\BIBentrySTDinterwordspacing}{\spaceskip=0pt\relax}
\providecommand{\BIBentryALTinterwordstretchfactor}{4}
\providecommand{\BIBentryALTinterwordspacing}{\spaceskip=\fontdimen2\font plus
\BIBentryALTinterwordstretchfactor\fontdimen3\font minus
  \fontdimen4\font\relax}
\providecommand{\BIBforeignlanguage}[2]{{%
\expandafter\ifx\csname l@#1\endcsname\relax
\typeout{** WARNING: IEEEtran.bst: No hyphenation pattern has been}%
\typeout{** loaded for the language `#1'. Using the pattern for}%
\typeout{** the default language instead.}%
\else
\language=\csname l@#1\endcsname
\fi
#2}}
\providecommand{\BIBdecl}{\relax}
\BIBdecl

\bibitem{kurosu2020human}
M.~Kurosu, \emph{Human-Computer Interaction. Multimodal and Natural
  Interaction: Thematic Area, HCI 2020, Held as Part of the 22nd International
  Conference, HCII 2020, Copenhagen, Denmark, July 19--24, 2020, Proceedings,
  Part II}.\hskip 1em plus 0.5em minus 0.4em\relax Springer Nature, 2020, vol.
  12182.

\bibitem{rasenberg2022multimodal}
M.~Rasenberg, W.~Pouw, A.~{\"O}zy{\"u}rek, and M.~Dingemanse, ``The multimodal
  nature of communicative efficiency in social interaction,'' \emph{Scientific
  Reports}, vol.~12, no.~1, p. 19111, 2022.

\bibitem{TURK2014189}
\BIBentryALTinterwordspacing
M.~Turk, ``Multimodal interaction: A review,'' \emph{Pattern Recognition
  Letters}, vol.~36, pp. 189--195, 2014. [Online]. Available:
  \url{https://www.sciencedirect.com/science/article/pii/S0167865513002584}
\BIBentrySTDinterwordspacing

\bibitem{oviatt2003advances}
S.~Oviatt, ``Advances in robust multimodal interface design,'' \emph{IEEE
  computer graphics and applications}, vol.~23, no.~05, pp. 62--68, 2003.

\bibitem{nayak2022improving}
P.~Nayak, T.~Higuchi, A.~Gupta, S.~Ranjan, S.~Shum, S.~Sigtia, E.~Marchi,
  V.~Lakshminarasimhan, M.~Cho, S.~Adya \emph{et~al.}, ``Improving voice
  trigger detection with metric learning,'' \emph{arXiv preprint
  arXiv:2204.02455}, 2022.

\bibitem{qin2021proximic}
Y.~Qin, C.~Yu, Z.~Li, M.~Zhong, Y.~Yan, and Y.~Shi, ``Proximic: Convenient
  voice activation via close-to-mic speech detected by a single microphone,''
  in \emph{Proceedings of the 2021 CHI Conference on Human Factors in Computing
  Systems}, 2021, pp. 1--12.

\bibitem{zhao2019raise}
S.~Zhao, B.~Westing, S.~Scully, H.~Nieto, R.~Holenstein, M.~Jeong, K.~Sridhar,
  B.~Newendorp, M.~Bastian, S.~Raman \emph{et~al.}, ``Raise to speak: An
  accurate, low-power detector for activating voice assistants on
  smartwatches,'' in \emph{Proceedings of the 25th ACM SIGKDD International
  Conference on Knowledge Discovery \& Data Mining}, 2019, pp. 2736--2744.

\bibitem{harel1987statecharts}
D.~Harel, ``Statecharts: A visual formalism for complex systems,''
  \emph{Science of computer programming}, vol.~8, no.~3, pp. 231--274, 1987.

\bibitem{chung2014empirical}
J.~Chung, C.~Gulcehre, K.~Cho, and Y.~Bengio, ``Empirical evaluation of gated
  recurrent neural networks on sequence modeling,'' \emph{arXiv preprint
  arXiv:1412.3555}, 2014.

\bibitem{1407896}
S.~Kettebekov, M.~Yeasin, and R.~Sharma, ``Prosody based audiovisual coanalysis
  for coverbal gesture recognition,'' \emph{IEEE Transactions on Multimedia},
  vol.~7, no.~2, pp. 234--242, 2005.

\bibitem{kettebekov2003improving}
------, ``Improving continuous gesture recognition with spoken prosody,'' in
  \emph{2003 IEEE Computer Society Conference on Computer Vision and Pattern
  Recognition, 2003. Proceedings.}, vol.~1.\hskip 1em plus 0.5em minus
  0.4em\relax IEEE, 2003, pp. I--I.

\bibitem{pouw2019quantifying}
W.~Pouw and J.~A. Dixon, ``Quantifying gesture-speech synchrony,'' in \emph{the
  6th gesture and speech in interaction conference}.\hskip 1em plus 0.5em minus
  0.4em\relax Universitaetsbibliothek Paderborn, 2019, pp. 75--80.

\bibitem{nagrani2021attention}
A.~Nagrani, S.~Yang, A.~Arnab, A.~Jansen, C.~Schmid, and C.~Sun, ``Attention
  bottlenecks for multimodal fusion,'' \emph{Advances in Neural Information
  Processing Systems}, vol.~34, pp. 14\,200--14\,213, 2021.

\bibitem{9190246}
K.~Gadzicki, R.~Khamsehashari, and C.~Zetzsche, ``Early vs late fusion in
  multimodal convolutional neural networks,'' in \emph{2020 IEEE 23rd
  International Conference on Information Fusion (FUSION)}, 2020, pp. 1--6.

\bibitem{Ramachandram2017DeepML}
D.~Ramachandram and G.~W. Taylor, ``Deep multimodal learning: A survey on
  recent advances and trends,'' \emph{IEEE Signal Processing Magazine},
  vol.~34, pp. 96--108, 2017.

\bibitem{summaira2021recent}
J.~Summaira, X.~Li, A.~M. Shoib, S.~Li, and J.~Abdul, ``Recent advances and
  trends in multimodal deep learning: A review,'' \emph{arXiv preprint
  arXiv:2105.11087}, 2021.

\bibitem{ceraso2014re}
S.~Ceraso, ``(re) educating the senses: Multimodal listening, bodily learning,
  and the composition of sonic experiences,'' \emph{College English}, vol.~77,
  no.~2, pp. 102--123, 2014.

\bibitem{dumas2009multimodal}
B.~Dumas, D.~Lalanne, and S.~Oviatt, ``Multimodal interfaces: A survey of
  principles, models and frameworks,'' \emph{Human machine interaction:
  Research results of the mmi program}, pp. 3--26, 2009.

\bibitem{holzapfel2004implementation}
H.~Holzapfel, K.~Nickel, and R.~Stiefelhagen, ``Implementation and evaluation
  of a constraint-based multimodal fusion system for speech and 3d pointing
  gestures,'' in \emph{Proceedings of the 6th international conference on
  Multimodal interfaces}, 2004, pp. 175--182.

\bibitem{williams2020understanding}
A.~S. Williams and F.~R. Ortega, ``Understanding gesture and speech multimodal
  interactions for manipulation tasks in augmented reality using unconstrained
  elicitation,'' \emph{Proceedings of the ACM on Human-Computer Interaction},
  vol.~4, no. ISS, pp. 1--21, 2020.

\bibitem{Qiu_2020}
\BIBentryALTinterwordspacing
X.~Qiu, Z.~Feng, X.~Yang, and J.~Tian, ``Multimodal fusion of speech and
  gesture recognition based on deep learning,'' \emph{Journal of Physics:
  Conference Series}, vol. 1453, no.~1, p. 012092, jan 2020. [Online].
  Available: \url{https://dx.doi.org/10.1088/1742-6596/1453/1/012092}
\BIBentrySTDinterwordspacing

\bibitem{brown2007shake2talk}
L.~M. Brown and J.~Williamson, ``Shake2talk: multimodal messaging for
  interpersonal communication,'' in \emph{Haptic and Audio Interaction Design:
  Second International Workshop, HAID 2007 Seoul, South Korea, November 29-30,
  2007 Proceedings 2}.\hskip 1em plus 0.5em minus 0.4em\relax Springer, 2007,
  pp. 44--55.

\bibitem{1389771}
R.~Stiefelhagen, C.~Fugen, R.~Gieselmann, H.~Holzapfel, K.~Nickel, and
  A.~Waibel, ``Natural human-robot interaction using speech, head pose and
  gestures,'' in \emph{2004 IEEE/RSJ International Conference on Intelligent
  Robots and Systems (IROS) (IEEE Cat. No.04CH37566)}, vol.~3, 2004, pp.
  2422--2427 vol.3.

\bibitem{7472168}
I.~Rodomagoulakis, N.~Kardaris, V.~Pitsikalis, E.~Mavroudi, A.~Katsamanis,
  A.~Tsiami, and P.~Maragos, ``Multimodal human action recognition in assistive
  human-robot interaction,'' in \emph{2016 IEEE International Conference on
  Acoustics, Speech and Signal Processing (ICASSP)}, 2016, pp. 2702--2706.

\bibitem{9812670}
M.~Saleh and I.~Jouny, ``Multimodal person identification through the fusion of
  face and voice biometrics,'' in \emph{2022 17th Annual System of Systems
  Engineering Conference (SOSE)}, 2022, pp. 164--169.

\bibitem{donkal2018multimodal}
G.~Donkal and G.~K. Verma, ``A multimodal fusion based framework to reinforce
  ids for securing big data environment using spark,'' \emph{Journal of
  information security and applications}, vol.~43, pp. 1--11, 2018.

\bibitem{7169562}
N.~Neverova, C.~Wolf, G.~Taylor, and F.~Nebout, ``Moddrop: Adaptive multi-modal
  gesture recognition,'' \emph{IEEE Transactions on Pattern Analysis and
  Machine Intelligence}, vol.~38, no.~8, pp. 1692--1706, 2016.

\bibitem{martin1997det}
A.~Martin, G.~Doddington, T.~Kamm, M.~Ordowski, and M.~Przybocki, ``The det
  curve in assessment of detection task performance,'' National Inst of
  Standards and Technology Gaithersburg MD, Tech. Rep., 1997.

\bibitem{kingma2014adam}
D.~P. Kingma and J.~Ba, ``Adam: A method for stochastic optimization,''
  \emph{arXiv preprint arXiv:1412.6980}, 2014.

\bibitem{horiguchi2019significance}
S.~Horiguchi, D.~Ikami, and K.~Aizawa, ``Significance of softmax-based features
  in comparison to distance metric learning-based features,'' \emph{IEEE
  transactions on pattern analysis and machine intelligence}, vol.~42, no.~5,
  pp. 1279--1285, 2019.

\bibitem{jia2021marblenet}
F.~Jia, S.~Majumdar, and B.~Ginsburg, ``Marblenet: Deep 1d time-channel
  separable convolutional neural network for voice activity detection,'' in
  \emph{ICASSP 2021-2021 IEEE International Conference on Acoustics, Speech and
  Signal Processing (ICASSP)}.\hskip 1em plus 0.5em minus 0.4em\relax IEEE,
  2021, pp. 6818--6822.

\bibitem{9751601}
Z.~Yu, Y.~Lu, Q.~An, C.~Chen, Y.~Li, and Y.~Wang, ``Real-time multiple gesture
  recognition: Application of a lightweight individualized 1d cnn model to an
  edge computing system,'' \emph{IEEE Transactions on Neural Systems and
  Rehabilitation Engineering}, vol.~30, pp. 990--998, 2022.

\bibitem{neverova2015moddrop}
N.~Neverova, C.~Wolf, G.~Taylor, and F.~Nebout, ``Moddrop: adaptive multi-modal
  gesture recognition,'' \emph{IEEE Transactions on Pattern Analysis and
  Machine Intelligence}, vol.~38, no.~8, pp. 1692--1706, 2015.

\bibitem{xu2022multimodal}
P.~Xu, X.~Zhu, and D.~A. Clifton, ``Multimodal learning with transformers: A
  survey,'' \emph{arXiv preprint arXiv:2206.06488}, 2022.

\end{thebibliography}

\end{document}